\documentclass[10pt,conference,compsocconf]{IEEEtran}

\ifCLASSOPTIONcompsoc
  \usepackage[nocompress]{cite}
\else
  \usepackage{cite}
\fi

\DeclareRobustCommand{\IEEEauthorrefmark}[1]{\smash{\textsuperscript{\footnotesize #1}}}

\usepackage{newtxtext,newtxmath}
\usepackage{microtype}
\usepackage{todonotes}
\usepackage{float}
\usepackage{mathtools}
\usepackage{graphicx}
\graphicspath{{figures}}
\DeclareGraphicsExtensions{.pdf,.png}
\usepackage{algorithmic}
\usepackage{url}
\usepackage{acronym}
\usepackage{tikz,nccmath}
\usepackage{cuted}  
\usepackage{caption,subcaption}

\usepackage{tabularray}
\UseTblrLibrary{siunitx,booktabs}

\usepackage[british]{babel}
\usepackage[english=british]{csquotes}
\usepackage{multirow}
\usepackage{xcolor}
\usetikzlibrary{decorations.pathmorphing}
\tikzset{snake it/.style={decorate, decoration=snake}}
\tikzstyle{snakeline} = [thick, decorate, decoration={pre length=2cm,
    post length=2cm, snake, amplitude=12mm,
    segment length=20mm}, magenta]

\usepackage{balance}
\usepackage{xspace}
\newcommand{\ie}{i.e.\@\xspace}
\newcommand{\eg}{e.g.\@\xspace}

\DeclareMathOperator*{\argmax}{arg\,max}
\usepackage[symbol]{footmisc}

\begin{document}
\newcounter{mytempeqncnt}

%
\title{%
    When the Ground Truth is not True:\\
    Modelling Human Biases in Temporal Annotations%
}

\author{
\IEEEauthorblockN{Taku Yamagata\IEEEauthorrefmark{1}\IEEEauthorrefmark{*}, Emma L. Tonkin\IEEEauthorrefmark{1}\IEEEauthorrefmark{*}, Benjamin Arana Sanchez\IEEEauthorrefmark{1}, Ian Craddock\IEEEauthorrefmark{1}, Miquel Perello Nieto\IEEEauthorrefmark{1}, \\ Raul Santos-Rodriguez\IEEEauthorrefmark{1},
Weisong Yang\IEEEauthorrefmark{1}, Peter Flach\IEEEauthorrefmark{1}}
\IEEEauthorblockA{\IEEEauthorrefmark{1}University of Bristol, United Kingdom\\
\{taku.yamagata,e.l.tonkin,benjamin.aranasanchez,ian.craddock,miquel.perellonieto,enrsr,ws.yang,peter.flach\}@bristol.ac.uk \\
}

}


%



\maketitle

\footnotetext[1]{These authors contributed equally to this work.}


\begin{abstract}
    \textls[-5]{In supervised learning, low quality annotations lead to poorly performing classification and detection models, while also rendering evaluation unreliable. This is particularly apparent on temporal data, where annotation quality is affected by multiple factors. For example, in the post-hoc self-reporting of daily activities, cognitive biases are one of the most common ingredients. In particular, reporting the start and duration of an activity after its finalisation may incorporate biases introduced by personal time perceptions, as well as the imprecision and lack of granularity due to time rounding. Here we propose a method to model human biases on temporal annotations and argue for the use of soft labels. Experimental results in synthetic data show that soft labels provide a better approximation of the ground truth for several metrics. We showcase the method on a real dataset of daily activities.}
\end{abstract}


%
\IEEEpeerreviewmaketitle

\section{Introduction}


\textls[-10]{The development of systems designed for detection and monitoring of everyday human activities (\eg movement, cooking, hygiene, sleep) benefits from accurate labelled (annotated) data. Diarising methods (participant self-reporting of events), often used for self-annotation, display several limitations. Reliability is limited, \eg Möller et al \cite{10.1145/2470654.2481406} found that subjects in a 6-week study of smartphone use reported at most only 70\% of detected events, perhaps due to forgetfulness. Intentional misreporting may also be a factor in recording of some activities, \eg fitness \cite{lester2006practical}. Since the contemporaneous annotation of a task usually requires interruption of a participant's activity, it is also likely that participants will not record events contemporaneously, to retain goal focus and avoid excessive task switching \cite{tonkin2018talk} and for practical reasons, \eg it is impractical to input or write down annotations during a task such as vacuuming or taking a shower. As a consequence, participant-contributed annotations may be characterised as unreliable and incomplete, in terms of \textit{event logging}, \textit{event start\slash end time} and \textit{event duration}.}

\textls[-10]{Uncertainty in the estimation of time reflects participant use of coarse temporal units to describe time and duration (\eg an event began around a certain time, estimated with a certain granularity [`around half-past'] and preference for `prototypical' times \cite{labiancaetal,mcgrath1988place} and is estimated with a certain duration). Self-reporting of event duration has been studied in linguistics (often via large corpora \cite{pan2007modeling}) and in music \cite{6587770}.  Precision of estimation of time of day reflects participant time awareness. Time perception is complex, elastic \cite{DROITVOLET2007504} and varies according to demographics such as age \cite{wittmann2005age} and context, \eg time of day \cite{CAMPBELL2001589}. It is also worth noting that disrupted time awareness is a clinical feature in various diagnoses, such as Alzheimer's disease \cite{requena2020altered}. At present, the characteristics of participant temporal annotation are not widely modelled in analysis of participant-labelled sensor data.} 

\textls[-10]{In summary, while a thorough discussion is beyond the scope of this paper, unless supported by electronic time-stamping, participants' timings for activity start, end and duration are often inexact; we give an example of this in Section 2. This quality issue may be compared with Kwon et al's `label jitter' \cite{10.1145/3341163.3347744}. However, jitter in post-hoc annotation (\eg of video) is likely to be of the order of fractions of a second. As we see in Section 2, participant estimation of activity times is found to have large uncertainty. Much of the work on both evaluating label quality \cite{hypothesis} and learning from weak labels \cite{9767420} originates from post-hoc annotation, in which multiple-annotator performance can be compared (\eg via statistical measures) and probability distributions built accordingly.}

This paper explores a means of modeling temporal uncertainty when making use of participant contributed labels describing time-series annotation tasks, and provides a first step in exploring the method's suitability for particular applications (\eg evaluation of predictions, training models); it is organised as follows. Section \ref{sec:case_study} presents a motivating case study. Section \ref{sec:method} describes our approach to estimating the annotation ambiguity and computing soft labels. Section \ref{sec:use_cases} demonstrates the characteristics of the soft labels with artificial and real-world datasets. Section \ref{sec:discussion} presents critical discussions of our approach. Finally, Section \ref{sec:conclusion} concludes with a summary and possible future work.


\section{Case study: SRM-17 recording of activities of daily living}
\label{sec:case_study}

\begin{figure}[!t]
    \centering
    \includegraphics[width=.95\linewidth]{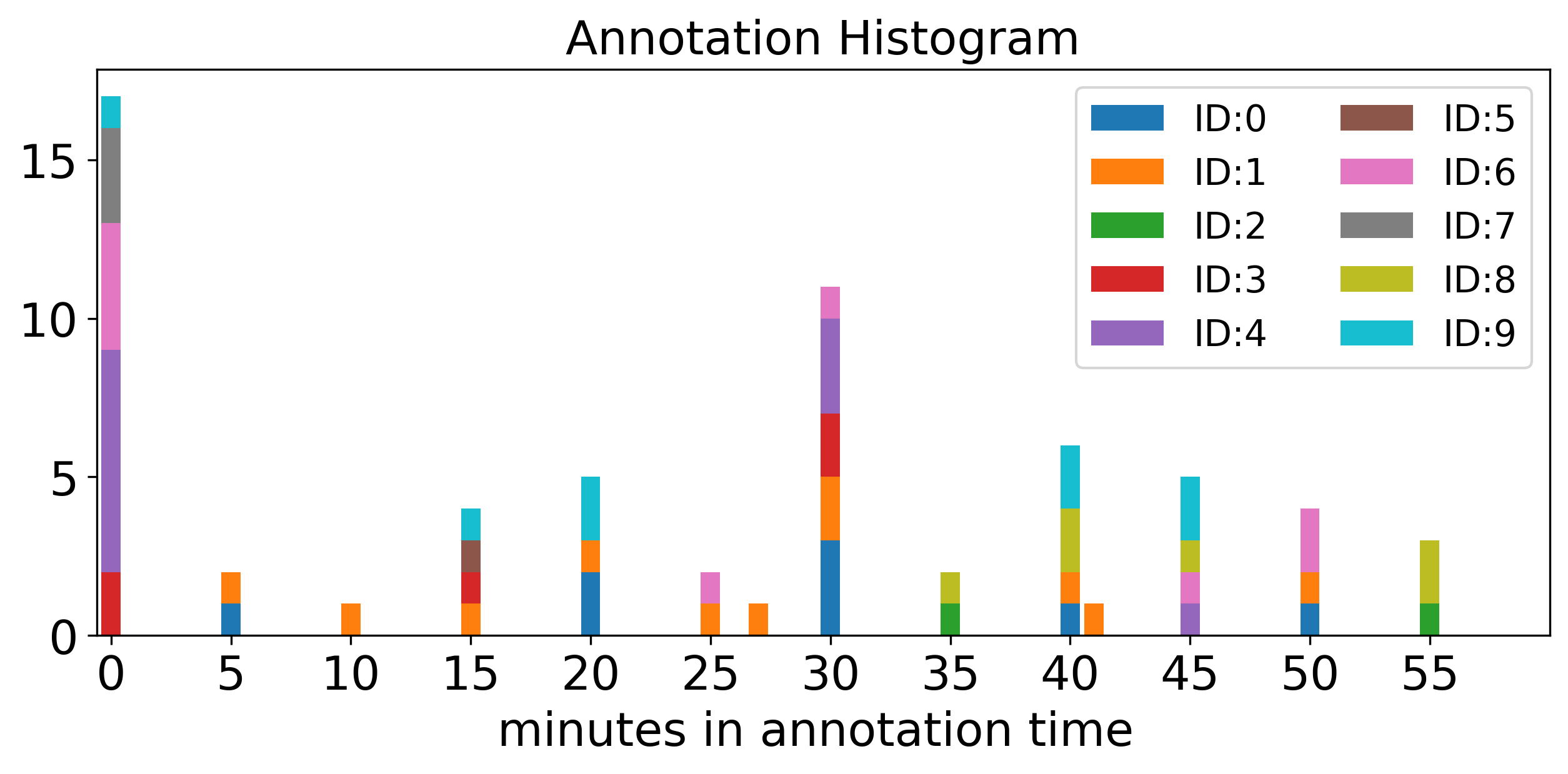}%
    \caption{Histogram of granularity patterns found on SRM-17 shower/bath annotations for dataset SPHERE 100-Homes, coloured by annotator.}
    \label{fig:annotations_hist}
\end{figure}

\textls[-10]{During the SPHERE 100-homes project, a subset of participants were encouraged to provide daily annotations via a modified Social Rhythm Metric (SRM-17), designed to measure daily lifestyle regularity \cite{monk2002simple}, self-administered by participants for up to seven days, resulting in approximately 100 days of data. Analysis of this small dataset demonstrated many features identified in the introduction. Taking the example of shower/bath events, as can be seen in Figure \ref{fig:annotations_hist}, participants' contributed annotations fall on particular times within the hour, \eg on the hour or half past.  
Data often appears to be reported to the nearest five minutes. This does not reflect instructions given to participants, who were given no specific guidance regarding precision of time reporting; researchers expected participants to report time read from a watch or clock. That said, in the clinical domain it is common for instructions to mandate a relatively low precision of time reporting. For example, the frequently-used Hauser home diary to assess functional status in Parkinson's disease asks participants to fill out diaries every half hour \cite{hauser2004parkinson}. Therefore, we may envisage several circumstances in which data analysts working with clinical data may find themselves in one of two conditions: annotation reporting is constrained by \textit{diary frequency} (\eg `during the last half hour'); or by \textit{participants' chosen granularity of time approximation} (\eg to the nearest five minutes, half hour, etc).}



\textls[-10]{While we would like to evaluate how well predictions made on the basis of sensor data correlate to the annotations provided by participants, sensor data itself may give inaccurate timings. Activity and event predictions rely on signals that vary due to incidental characteristics about a home \eg measurement of heat from cooking varies by stove type, placement, response time and sensor placement\cite{COFFEY2021100065}. Hence,\cite{COFFEY2021100065} apply a `thermal lag' parameter to account for the time it takes for a temperature increase to be detected. 
Similarly, humidity sensors used to predict when shower or bath events occurred exhibit a `humidity lag', due to characteristics of the bathroom appliances (\eg shower, bath), configuration (\eg closed shower stall, curtain, no enclosure from the bathroom), and architecture (\eg room size, window placement and ventilation such as extractor fans).}

%

\section{Method}
\label{sec:method}

We propose to tackle the uncertainty in time-series annotations in two stages. In the first step, we predict the time resolution each annotator uses (ambiguity) with a Bayesian approach. In the second stage, we produce soft labels based on the predicted annotation's time resolution. It is also important to consider how to evaluate a target model with soft labels (\ie how to use the soft labels), but this is outside the current paper's scope.

\begin{figure}[ht]
    \centering
    \includegraphics[width=1.0\linewidth]{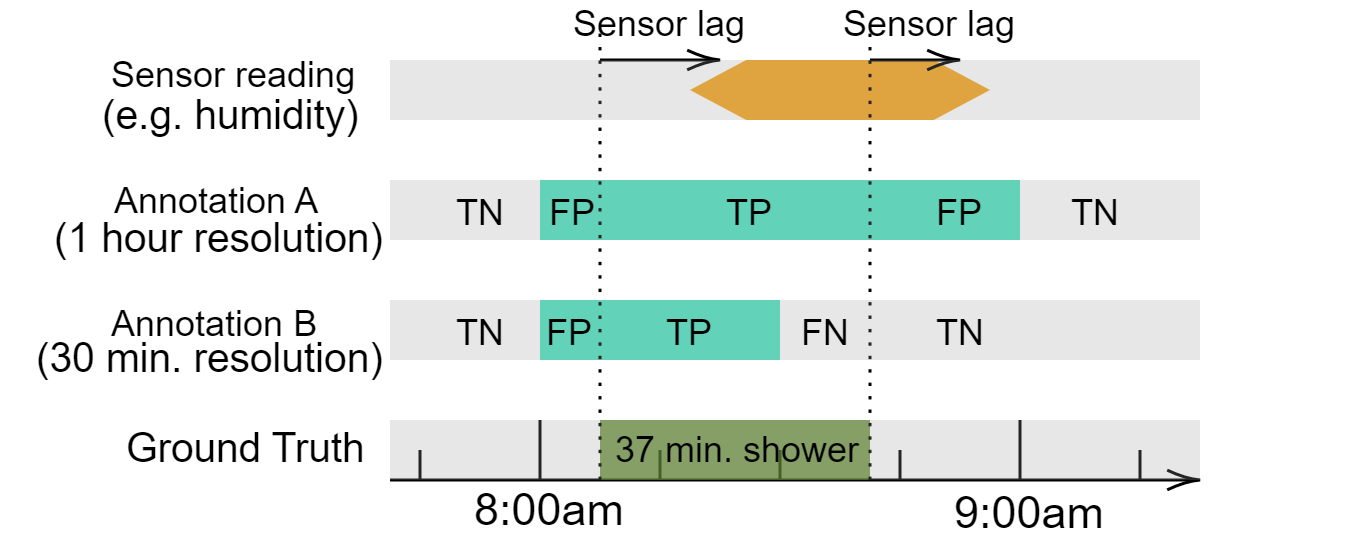}
    \caption{\textls[-5]{Examples of biases added during the annotation process. The bottom graph shows the time in which the person took a 37 minutes shower. The graph above shows a possible annotation \textbf{B} estimating that it took about 30 minutes at 8am, while the graph above that shows a different annotation \textbf{A} estimating a shower of an hour at 8am. Each annotation indicates the true (and false) positives (and negatives) that would result from comparison with the ground truth.}}
\end{figure}

\subsection{Predicting Annotator's Habit and Annotation's Uncertainties}

First, we introduce predefined categories (\(c_n\)) for annotation time resolutions. Each category has a subset of possible annotation values ($\mathbb{C}_n$) that indicate how the person who belongs to the category tends to annotate. For example, if we define the 1st category ($c_1$) as 30 minutes time resolution, then the possible annotation times would be 0 and 30 minutes ($\mathbb{C}_1=\{0, 30\}$). We use the following five categories to demonstrate our method throughout this paper.

\begin{tabular}{lll}
  $c_1$ & : 30 mins. res. & $\mathbb{C}_1=\{0, 30\}$ \\
  $c_2$ & : 15 mins. res. & $\mathbb{C}_2=\{0, 15, 30, 45\}$\\
  $c_3$ & : 10 mins. res. & $\mathbb{C}_3=\{0, 10, 20, 30, 40, 50\}$\\
  $c_4$ & :  5 mins. res.  & $\mathbb{C}_4=\{0, 5, 10, 15, \dots , 55\}$\\
  $c_5$ & :  1 mins. res.  & $\mathbb{C}_5=\{0, 1, 2, 3, 4, \dots , 59\}$\\
\end{tabular}

\textls[-10]{Then, we introduce two random variables. One indicates the annotator's habit ($H_j$), which indicates which time resolution category the annotator tends to use, and the other is a category ($C_{j,i}$) that is actually used for each annotation. The index $j$ and $i$ are the annotators and the annotations index, respectively. $C_{j,i}$ is defined for each annotation, and $H_j$ is defined for each annotator. Now, we introduce our method to derive the posterior probabilities of the two random variables ($H_j$ and $C_{j,i}$). For simplicity, we drop the annotator index $j$ in the rest of the paper, as the following process can be repeated for each annotator separately. We present the notation below.}

\begin{tabular}{lp{0.62\linewidth}} 
  $\mathcal{D}=\{d_i\}_{i=1}^{N}$ & Set of annotations \\
  $N$ & Number of annotations \\
  $d^*_i$ & True value for an annotation $d_i$ \\
  $H$ & Annotator's habit \\
  $C_i$ & i-th annotation category \\
  $c_n$ & Categories for $H$ and $C_i$ \\
  $|c_n|$ & Number of annotations in category $c_n$ \\
  $T_{c_n}$ & Time resolution for category $c_n$ \\
  $N_c$ & Number of the categories
\end{tabular}

\begin{figure}[!t]
    \centering
    \includegraphics[width=0.9\linewidth]{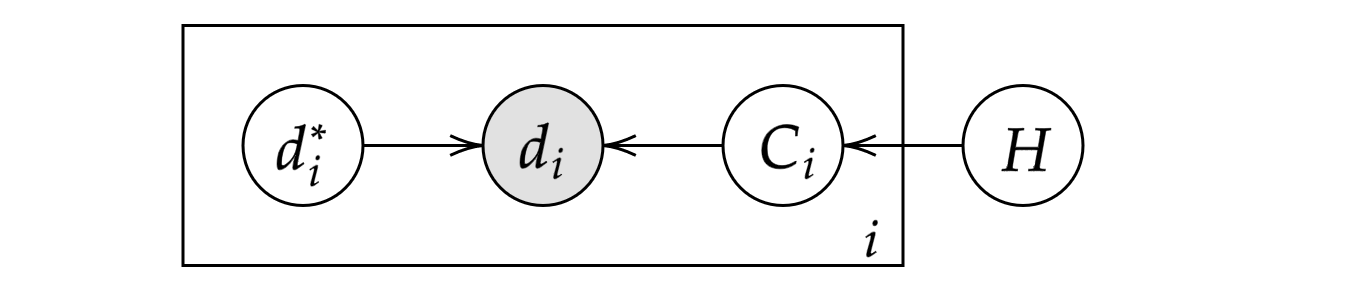}%
    \caption{\textls[-10]{Probabilistic graphical model relating the variables for annotations. We assume that the annotation $d_{i}$ is generated from the ground truth $d^*_{i}$ and the annotation time resolution $C_i$. The annotation time resolution depends upon the annotator's habit $H$. The index $i$ is the annotation's index.}}
    \label{fig:annotators_graph_model}
\end{figure}

\textls[-10]{We assume that the $i$-th annotations ($d_{i}$) are generated from $C_{i}$ and $d^*_i$, and $C_{i}$ is generated from $H$. Figure~\ref{fig:annotators_graph_model} shows the probabilistic graphical model relating the variables. To infer the annotator's $H$ and annotations' $C_i$, we compute their posterior probabilities given annotations $\mathcal{D}$ \ie $P(H|\mathcal{D})$ and $P(C_i|\mathcal{D})$.}

\begin{equation}
    \begin{split}
    \label{eq:p_H_D}
        P(H|\mathcal{D}) = \medmath{\frac{P(H)}{P(\mathcal{D})} \prod_{i=1}^{N} \sum_{C_i} P(C_i|H) P(d_i|C_i)} \\
    \end{split}
\end{equation}
Eq.~\ref{eq:p_H_D} shows how we can compute $P(H|\mathcal{D})$. 
Its detailed derivations are in Appendix A. 
The term $P(d_i|C_i)$ is the probability of getting the annotation $d_i$ with the given annotation category $C_i$. Here, we assume a uniform distribution for $P(d^*_i)$; hence we define it as Eq.~\ref{eq:p_di_ci}.
\begin{equation} \label{eq:p_di_ci}
    \medmath{P(d_i|C_i) = 
    \begin{cases}
        \frac{1}{|C_i|}, & \text{if } d_i \in \mathbb{C}_i \\
        0,               & \text{otherwise}
    \end{cases}}
\end{equation}
where $|C_i|$ is a number of annotations in the category $C_i$. For example, when $C_i=c_1$, $|C_i|=2$ as $c_1$ has two members $\mathbb{C}_1=\{0, 30\}$. Hence the probability of $d_i=0$ or $30$ when $C_i=c1$ is $0.5$ in Eq.~\ref{eq:p_di_ci}. The term $P(C_i|H)$ is the probability of using the annotation category $C_i$ when the annotator's habit is $H$. We use the following $P(C_i|H)$ in this work:
\begin{equation} \label{eq:p_ci_H}
    P(C_i|H) = 
    \begin{cases}
        1-\delta,               & \text{if } C_i = H \\
        \frac{\delta}{N_c-1},   & \text{otherwise}
    \end{cases}
\end{equation}
where $\delta \in [0,1]$ is the probability of taking a different annotation category to the annotator's habit $H$. In this paper, we use $\delta=0.1$. Further discussion regarding the choice of $\delta$ value is in Section~\ref{sec:discussion}. We assume the prior of $H$ is the uniform distribution, as we do not assume any prior knowledge about the annotator's habit. Now, we compute the posterior of $C_i$, which can be derived as follows (again, the detailed derivations are in Appendix B):
\begin{equation} \label{eq:p_ci_D}
\begin{split}
    P(C_i|\mathcal{D}) = \sum_{H} \frac{P(d_i|C_i) P(C_i|H)}{P(d_i|H)} P(H|\mathcal{D}), \\
\end{split}
\end{equation}
\textls[-15]{where we can compute $P(d_i|C_i)$ from Eq.~\ref{eq:p_di_ci}, $P(C_i|H)$ from Eq.~\ref{eq:p_ci_H}, $P(H|D)$ from Eq.~\ref{eq:p_H_D} and $P(d_i|H)=\sum_{C_i} P(d_i|C_i)P(C_i|H)$. Once we have computed the posteriors, we use them to produce the soft labels. It is possible to prepare the soft labels for all categories and combine them according to the posteriors (fully Bayesian approach). This work takes the \ac{MAP} estimation for $C_i$ and generates the soft label from it.}

\subsection{Generating Soft Labels}

\textls[-10]{The annotations for time-series data have been produced by recording an event's start and end time. 
We introduce a probability distribution over the timings based on the category $C_i$ detected in the previous subsection and compute soft labels that indicate a probability of the event happening at a given time. We adopt a uniform distribution for the true start and end timings $d^*_s$ and $d^*_e$ because we assume the annotations are generated by rounding the actual time. (Again, it is interesting for future work to study different distributions.) The distributions are uniform across the period of the inferred annotation interval and placed its centre at the annotated timing. 
For example, if a start timing annotation category $C_s$ is $c_3$, it becomes uniform distribution bounded by $d_s - 5$ and $d_s + 5$ as the category $c_3$ has annotations with 10 minutes intervals, where $d_s$ and $d_e$ are annotations for the start and end timing.}
\begin{equation} \label{eq:p_ds}
\begin{split}
    p(d^*_s) &= \mathrm{U}(d_s - T_{C_s}/2, d_s + T_{C_s}/2),  \\
    p(d^*_e) &= \mathrm{U}(d_e - T_{C_e}/2, d_e + T_{C_e}/2),  \\
\end{split}
\end{equation}
\textls[-10]{where $T_{C_s}$ and $T_{C_e}$ are the annotation time intervals for the inferred annotation categories $C_s$ and $C_e$.\footnote{These probabilities must be conditioned on $C_i$ and $d_i$ like $p(d^*_s|C_s, d_s)$. We drop the conditioning for simplicity here.} Then, we compute probabilities of the event that has started $P(d^*_s\leq t)$ and has not yet ended $P(d^*_e>t)$ at time $t$ by taking the integration of Eq.~\ref{eq:p_ds}.}

\begin{equation} \label{eq:p_event}
\begin{split}
    P(d^*_s\leq t) &= \int_{-\infty}^{t}p(d^*_s)~dd^*_s,  \\
    P(d^*_e>t) &= \int_{t}^{+\infty}p(d^*_e)~dd^*_e.  \\
\end{split}
\end{equation}
Finally, we compute the soft label (\(P^{\text{(label)}}(t)\)), which is the probability of the event at time \(t\) by multiplying $P(d^*_s<t)$ and $P(d^*_e>t)$. Here, we assume the start and end timings are statistically independent. It simplifies the following derivations and helps convey the idea of our approach. It is a strong assumption, and it does not hold in some cases. For example, if the start and end timings are close relative to the annotation time resolution, then we need to consider the dependency between the start and end timings -- the end timing must be later than the start timing. We want to extend our model to support such a scenario in the future.

\begin{equation} \label{eq:p_label}
\begin{split}
    P^{\text{(label)}}(t) &= P(d^*_s\leq t \wedge d^*_e>t) \\
                   &= P(d^*_s\leq t) P(d^*_e>t).\\
\end{split}
\end{equation}

\begin{figure}
    \centering
    \includegraphics[width = .95\linewidth]{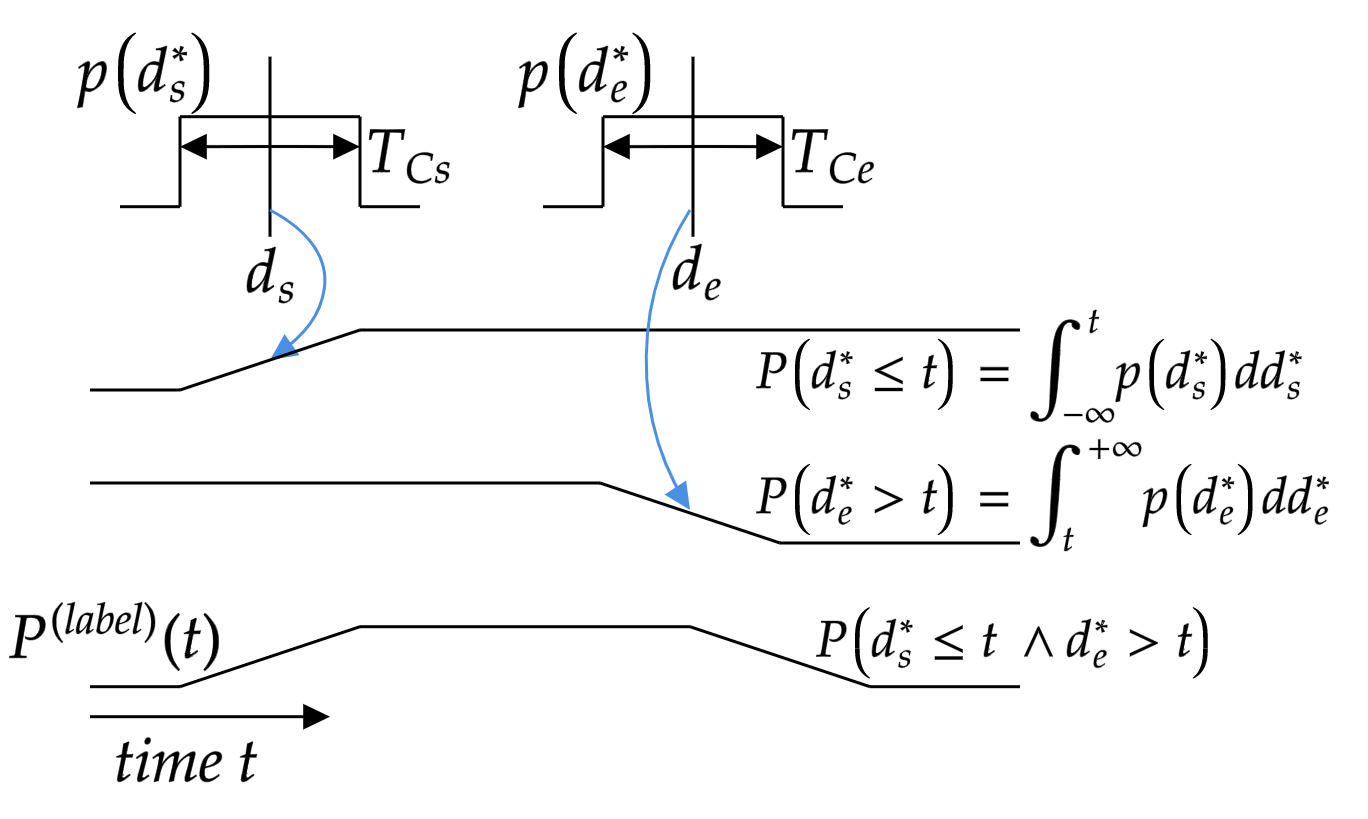}
    \caption{Soft label for time series data. Given start\slash end time annotations ($d_s$ and $d_e$) and the estimated time resolutions, we assume the probability distribution of the actual start\slash end time ($d^*_s$ and $d^*_e$) -- top of the figure. Then, we compute the probabilities that the event has already started and ended for each time slot. Finally, we compute the probability that the event has started but not ended for each time slot, becoming the soft label.}
    \label{fig:soft_label}
\end{figure}

\textls[-10]{Figure~\ref{fig:soft_label} shows the above soft label computation process with uniform distributions. It starts with the given start\slash end time annotations ($d_s$ and $d_e$) and the estimated time resolutions, we assume the probability distribution of the actual start\slash end time ($d^*_s$ and $d^*_e$). Then, compute the probabilities that the event has already started and ended for each time slot. Finally, we compute the probability that the event has started but not ended for each time slot, becoming the soft label.}

\section{Experiments}
\label{sec:use_cases}

In this section, we compare the hard and soft labels with artificial examples and real-world datasets. The hard labels have either true or false values and are set to true for the time slots between the start and end time annotations. The artificial examples allow us to demonstrate the characteristics of the soft labels in a controlled environment. We show the actual use case with real-world datasets. 

\subsection{Simple Task}

\begin{figure}[ht]
    \centering
    \includegraphics[width=0.9\linewidth]{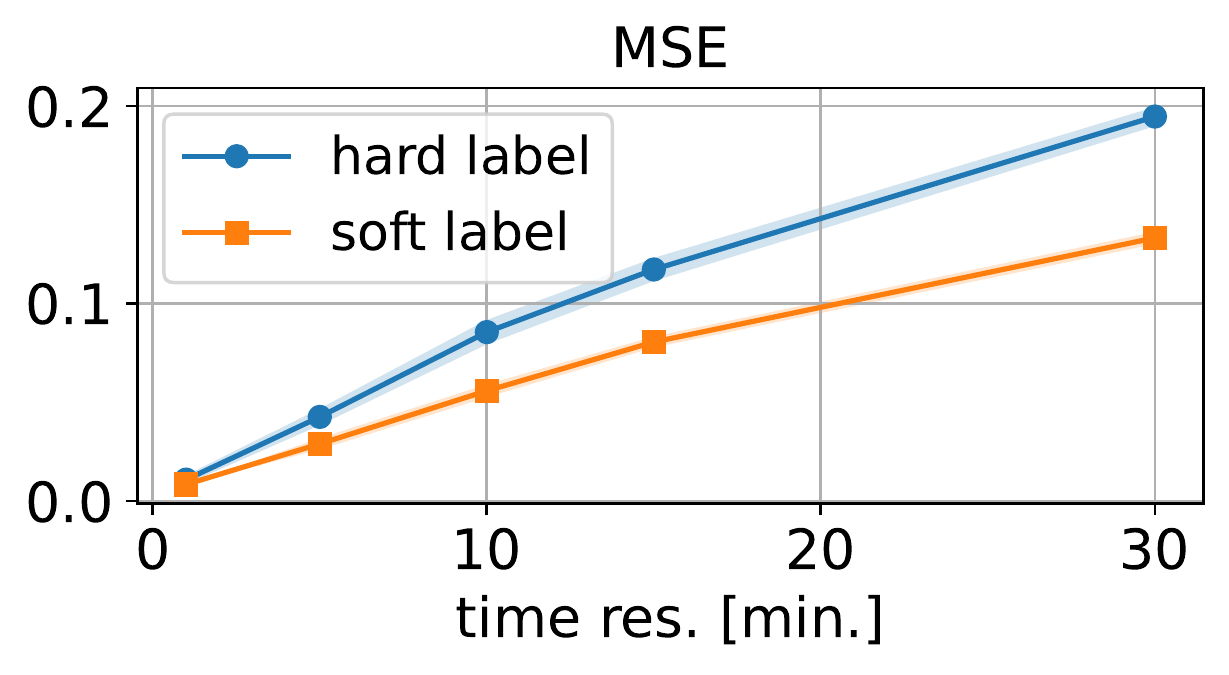}%
    \caption{MSE for the hard and soft labels against the ground truth. The X-axis is the annotation time resolution. The MSE measured around the ground truth start and end time \num{\pm 15} minutes range. The results show the soft label is better (smaller MSE) than the hard label.}
    \label{fig:mse}
\end{figure}

\textls[-10]{First, we compare the soft and conventional hard labels in a simple synthetic example. It consists in a series of tasks with a start and end events (\eg sleeping, cooking or showering). Here, we randomly generate the true start and end times (ground truth), then we make the annotations by rounding the times based on a given time resolution (\eg 5, 10, 15 and 30 minutes). We consider the rounded start and end times as the provided hard labels, and apply the proposed method to obtain the soft labels as described in Sec.~\ref{sec:method}. Finally, we evaluate the hard and soft labels by comparing them to the ground truth. Figure~\ref{fig:mse} shows the mean squared error (MSE) of the ground truth with respect to the hard and soft labels, with the time resolution of the annotations in the X-axis. The MSEs are measured around the ground truth start and end time \num{\pm 15} minutes range. The result suggests that the hard label has a progressively larger error than the soft label when the resolution interval of the time annotations increase. This clearly indicates that the proposed soft labels are less penalizing than the hard labels.}

\begin{figure}[t]
    \centering
    \includegraphics[width=0.9\linewidth]{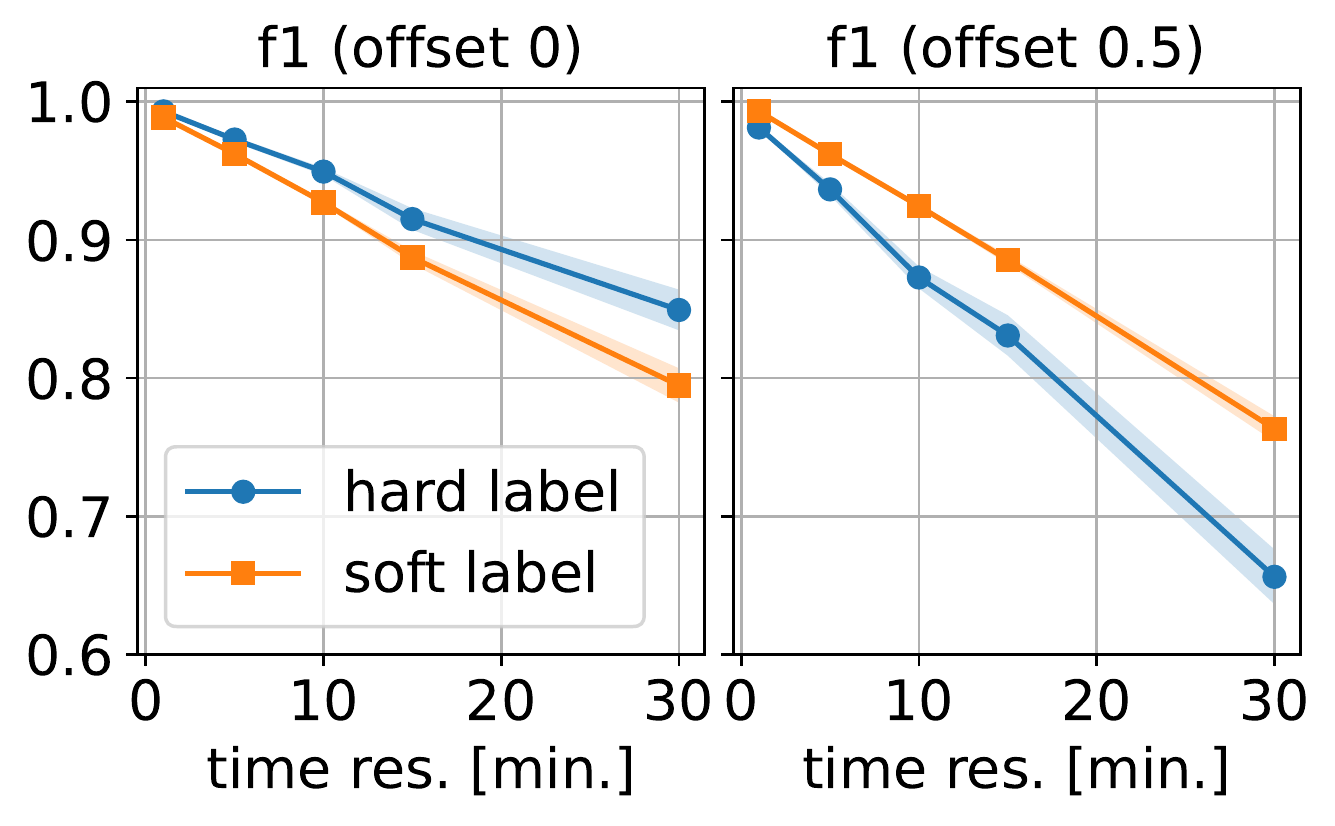}%
    \caption{\textls[-10]{F1 score for the hard and soft labels against the ground truth with different annotation time resolutions. The annotations are produced from the ground truth by applying a bias and rounding. Here, we use zero and half of the annotation time resolution as the bias (offset). The results suggest that the soft label performs better than the hard label with the bias.}}
    \label{fig:F1_0_0.5}
\end{figure}

\textls[-10]{Next, we perform a similar analysis with F1 score, as in most detection scenarios the positive class is more important than the negative one. The left plot of Figure~\ref{fig:F1_0_0.5} shows the results with the annotation time resolution on the X-axis. This suggests the hard label is better (higher F1 score) than the soft label. This is the opposite result to the MSE result and counter-intuitive as the soft label accurately reflects the degree of ambiguity by having values between zero and one, whereas the hard label has only either zero or one. The F1 score (or any metric based on the confusion matrix) is penalised by having a value between zero and one (like soft labels).}
Also, this is the best-case scenario for the hard label, as the annotation is produced by using rounding. Hence, the expected ground truth is matched with the hard label. We are interested in exploring evaluation metrics that do not penalise the prediction or ground truth ambiguity. However, this is left for future work.

\textls[-10]{We also evaluate the annotations produced by applying a bias (offset) and rounding to the ground truth. This is useful to simulate a case in which the annotator does not remember accurately the beginning and end of the event, which adds a possible bias on top of the rounding error. With this bias, the expected ground truth no longer matches the hard label. We set the bias to half of the annotation resolution and measure the F1 score for the hard and soft labels. The result (the right plot of Figure~\ref{fig:F1_0_0.5}) suggests that the soft label is better than the hard label in this case. We can see that the hard label results are degraded due to the bias in the annotation, and the soft label results stay the same as before. This is because we shift the uniform distribution for producing the soft label (Eq.~\ref{eq:p_ds}) based on the estimated annotation resolution.}


\subsection{Humidity Event Detection}

\textls[-10]{We test the soft and hard labels for evaluating a shower event detection model. The model uses the humidity sensor reading and predicts if someone is taking a shower at each time slot -- binary classification task. The model used is a hidden Markov model. It takes the humidity level as the observations and treats the shower status (on\slash off) as the hidden variable. We assume that the predictive model is already provided, as the learning process is not the focus of this paper; but the evaluation.}

\begin{table}[ht]
\caption{Confusion matrix for the shower event detection model with soft and hard labels.}
\label{tab:cm}
\begin{subtable}[c]{0.48\linewidth}
    \caption{With hard labels}
    \begin{tblr}{
            colspec={c *2{X[c]}},
            row{1} = {font = \bfseries, b},
            cell{2}{1} = {font = \bfseries},
            cell{1}{2} = {c=2}{c},
        }
        \toprule
         & Prediction & \\
        \cmidrule[lr]{2-3}
        Label & No & Yes \\
        \midrule
        No  & 3186 & 63 \\
        Yes & 41   & 70 \\
        \bottomrule
    \end{tblr}
\end{subtable}
\hfill
\begin{subtable}[c]{0.48\linewidth}
    \caption{With soft labels}
    \begin{tblr}{
            colspec={c *2{X[c]}},
            row{1} = {font = \bfseries, b},
            cell{2}{1} = {font = \bfseries},
            cell{1}{2} = {c=2}{c},
        }
        \toprule
         & Prediction & \\
        \cmidrule[lr]{2-3}
        Label & No & Yes \\
        \midrule
        No  & 3185.14 & 69.50 \\
        Yes & 41.86   & 63.50 \\
        \bottomrule
    \end{tblr}
\end{subtable}
\end{table}

\textls[-10]{We use the SRM-17 dataset~\cite{monk2002simple} with self-reported annotations. We pick the dataset for ID:4 (Figure~\ref{fig:annotations_hist}) as it has a coarse annotation time resolution. First, we generate the hard and soft labels based on self-report annotations and then compare them against the prediction model output. Table~\ref{tab:cm} shows the confusion matrices for the evaluation results with the hard and soft labels. We do not know the ground truth; hence we cannot directly compare the performance of these labels. We can say that the soft and hard labels give different results. We discuss approaches to further detailed characterisation and evaluation in the future work section of this paper.
We can also see that the results with the soft label are slightly worse than those with the hard labels (slightly fewer true-positives and true-negatives and more false-positives and false-negatives). It is because these matrices penalise the ambiguities (soft labels); as in Section 4.1, the suitability of these metrics for comparative evaluation is left for further discussion.}

\section{Discussion}
\label{sec:discussion}

In this section, we first discuss the findings from synthetic data and then present implications and limitations for real-world examples. 

\subsubsection*{Accuracy of the category estimation}

\textls[-10]{Our method estimates the posterior of the random variables $H$ and $C_i$. It is designed to pick the annotation category for coarser time resolution categories than the finer resolution ones. It is intuitively correct because if the annotations have only 30 minutes of the resolution, it is natural to think the annotator uses 30 minutes rather than the finer resolution that also possible to have 0 and 30 minutes, such as 15 minutes or 10 minutes. Our method wrongly estimates the annotation categories if all (or most) annotations landed on a coarser resolution time. The likelihood of such an error diminishes if it has more annotations.}

\begin{figure}[ht]
    \centering
    \includegraphics[width=0.9\linewidth]{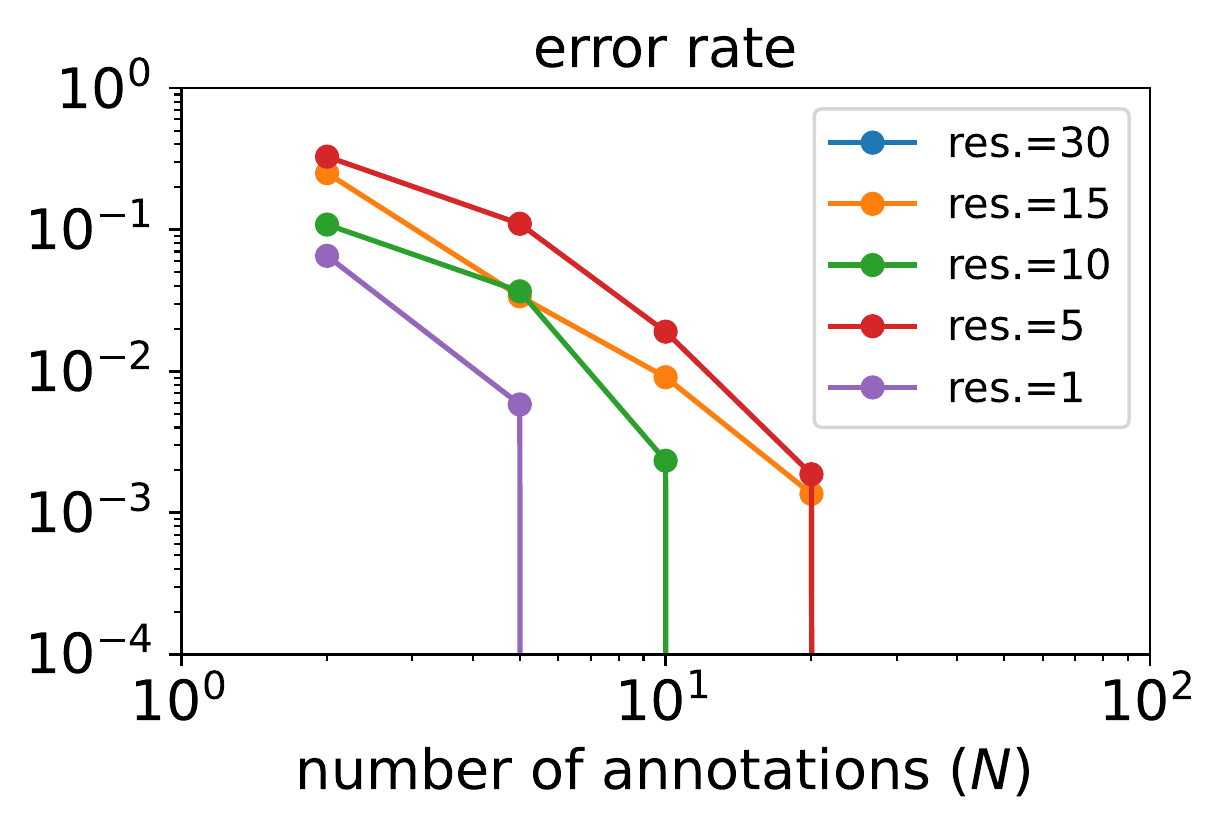}
    \caption{\textls[-15]{Error rate of the annotation's category estimation. The error rates are measured on synthetically generated annotations. Each line is for the annotation category. The X-axis is the number of annotations. The error rate decrease as it has more annotations.}}
    \label{fig:err_rate}
\end{figure}

Figure~\ref{fig:err_rate} shows the error rate of the annotation category estimation $\argmax_{C_i}{P(C_i|\mathcal{D})}$. As the plot suggests, the error rate decreases quickly with the more annotation it receives. Here we assume all annotations come from a single time resolution category. Each line indicates the annotation category for the annotation. There is no line for resolution=30 minutes as the error rate is always zero.

\subsubsection*{Choice of $\delta$ in Eq.~\ref{eq:p_ci_H}}

\textls[-10]{The parameter $\delta$ indicates how likely the annotator uses the different annotation categories from the habit ($H$). $\delta=0.1$ means that the annotator uses different annotation categories once in ten times. It affects both of the two random variable posteriors $P(H|\mathcal{D})$ and $P(C_i|\mathcal{D})$. If $\delta$ is small (close to zero), then $H$ would be the category compatible with all annotations. For example, if all but one annotation are 30 minutes (\eg 9:30), and one has 1-minute time resolution (\eg 13:16), then $H$ would be $c_5$ (1-minute resolution) when $\delta$ is too small.  
On the other hand, if $\delta$ is too large ($\sim 1.0$), then the posterior of $C_i$ might ignore the $H$ and pick the category based only on the annotation value $d_i$. For example, if all annotations are 5 minutes time resolution, then $H$ would correctly be $c_4$ (5-minutes resolution). However, $C_i$ would be wrongly $c_1$ if $d_i \in \mathbb{C}_1$ with too large $\delta$. We pick $\delta=0.1$ as it seems a good balance for picking the right $H$ and $C_i$.}

\subsubsection*{Choice of $p(d^*_i|d_i, C_i)$}

In this paper, we use the uniform distribution for $p(d^*_i|d_i, C_i)$ (Eq.~\ref{eq:p_di_ci}) as we assume a simple annotator behaviour -- just rounding the actual time. However, it is more plausible to use gradually increasing and decreasing distributions -- \eg Gaussian or trapezoidal shape distributions. This is left for future work. 

\subsubsection*{Learning from soft labels}

\textls[-10]{Our proposed method can be potentially applied during the learning process of a classifier\slash regressor as well.
For example, by setting the start and end of the events with the proposed soft labels, we can interpret the augmented regions as weak labels
\cite{Hernandez-Gonzalez2016, Perello2017, 9767420}.
Two possible algorithms that can be potentially used in this scenario are pseudo-labeling \cite{Dong2013} and Optimistic Superset Learning \cite{hullermeier2015}, which are iterative learning methods that consider the model's predictions among the candidate labels as correct, and retrain the model with those labels.
It is also possible to add those samples for which the model is more confident (e.g. exceeding a certain threshold).}




We could also consider the proposed soft labels as probabilities coming from a Bernoulli distribution, or as prior beliefs of belonging to the positive class.
Some algorithms that use similar soft labels are label smoothing \cite{Muller2019} and an Expectation Maximization method proposed by Jin and Ghahramani \cite{Jin2002}. 

\subsubsection*{Evaluation metrics for soft labelling}

\textls[-0]{The commonly used evaluation metrics for classification tasks (such as accuracy, precision, recall and F1 score) are penalised by the ambiguity in the labels or the predictions (the ambiguity reduces these scores). In this paper, we show that MSE does not have such an issue and correctly show the benefit of the soft labels in our experiment. Also, the MSE is similar to the Brier score, which is commonly used to evaluate probabilistic predictions' accuracy. It may suggest that the MSE is the right evaluation metric for soft labels. However, we need further study to understand which metric is appropriate in which scenario (objective) with the soft labels.}

\subsubsection*{Towards real-world comparison of soft and hard labels}

\textls[-10]{Real-world comparison of soft and hard labels in contemporaneous self-annotation by participants is complicated by the lack of wholly reliable ground truth, particularly in environments in which it is not possible to rely on data collection suitable for post-hoc annotation methods (\eg video is unlikely to be appropriate in domestic bathrooms). However, we may look toward other data sources to resolve or reduce some of the ambiguities identified. For example, we might look to sensor fusion, referencing other sensors for associated information such as presence, temperature or power use \eg to detect operating times of the appliance, or to get accurate bounds of a person's entrance into and departure from the room. This helps to resolve the confounding question of sensor lag (\eg time taken before sensor detects temperature or humidity rises), giving us a greater insight into real-world timings.}

\subsubsection*{Human performance in temporal estimation}

\textls[-10]{Time estimation performance and bias is a complex topic and the underlying mechanisms and their causes are largely beyond the scope of this paper. However, we hope that discussion of this mechanism may spark further exploration of the characterisation and handling of this aspect of self-reported data. }

\section{Conclusion}
\label{sec:conclusion}

\balance
We proposed a method to identify the annotator's approach to the task and the ambiguity that comes with it. Also, we devised a way to generate soft labels based on the estimated ambiguity. Our evaluation results suggest that the soft label is better in mean squared error than the hard label. However, the soft labels show worse results than the hard labels in terms of F1 score, because metrics like F1 score inherently penalise the ambiguities. We consider many avenues for future work, namely, improving the model of human annotations, designing new evaluation metrics for soft labels and using soft labels for the learning stage. We also hope to report on variance in granularity of participant annotation of other types of events in a future publication.

\ifCLASSOPTIONcompsoc
 \section*{Acknowledgments}
\else
 \section*{Acknowledgment}
\fi
This work was supported by the SPHERE Next Steps Project funded by EPSRC
under Grant EP/R005273/1. RSR is funded by the UKRI Turing AI Fellowship EP/V024817/1.

\begin{figure*}[!t]
\setcounter{mytempeqncnt}{\value{equation}}
\begin{equation} 
\label{eq:p_H_D_deriv}
\begin{split}
    P(H|\mathcal{D}) &= P(\mathcal{D}|H)P(H)/P(\mathcal{D}) \\
                     &= \sum_{C_1,\dots,C_{N}} \sum_{d^*_1,\dots,d^*_{N}} P(d_1, \cdots, d_{N}|d^*_1, \cdots, d^*_{N}, C_1, \cdots, C_{N}) P(C_1, \cdots, C_{N}|H) P(d^*_1, \cdots, d^*_{N}) \frac{P(H)}{P(\mathcal{D})} \\
                     &= \medmath{\sum_{C_1,\dots,C_{N}} \sum_{d^*_1,\cdots,d^*_{N}} \prod_{i=1}^{N} P(d_i|d^*_i, C_i) P(d^*_i) P(C_i|H) \frac{P(d^*_1, \cdots, d^*_{N})}{P(d^*_1)\cdots P(d^*_{N})} \frac{P(H)}{P(\mathcal{D})}} \\
                     &= \medmath{\frac{P(d^*_1, \cdots, d^*_{N})}{P(d^*_1)\cdots P(d^*_{N})} \frac{P(H)}{P(\mathcal{D})} \prod_{i=1}^{N} \sum_{C_i} P(C_i|H) \sum_{d^*_i} P(d_i|d^*_i, C_i) P(d^*_i)} \\
                     &= \medmath{\frac{P(d^*_1, \cdots, d^*_{N})}{P(d^*_1)\cdots P(d^*_{N})} \frac{P(H)}{P(\mathcal{D})} \prod_{i=1}^{N} \sum_{C_i} P(C_i|H) P(d_i|C_i)}
\end{split}
\end{equation}
\setcounter{equation}{\value{mytempeqncnt}+1}
\hrulefill
\vspace*{4pt}
\end{figure*}



\bibliographystyle{IEEEtran}
\textls[-15]{\bibliography{bibliography}}
%


\appendices

\section{$H$ posterior ($P(H|\mathcal{D})$) derivation}

Eq.~\ref{eq:p_H_D_deriv} shows how to derive \(P(H|\mathcal{D})\). We insert two sets of latent variables (\(C_1,\cdots, C_{N}\) and \(d^*_1,\cdots,d^*_{N}\)) and apply the chain rule to get the second line. Then, factorising $d_i$ and $C_i$ by using the graphical model shown in Figure~\ref{fig:annotators_graph_model}. Finally, the last line is derived by pushing the \(\sum\) operators into the relevant terms and replacing the term \(\sum_{d^*_i}P(d_i|d^*_i,C_i)d(d^*_i)\) with \(P(d_i|C_i)\). We also assume \(\frac{P(d^*_1, \cdots, d^*_{N})}{P(d^*_1)\cdots P(d^*_{N})} = 1.0\) (assumed \(d^*\) are all independent each other) to get Eq.~\ref{eq:p_H_D}.

\section{$C_i$ posterior ($P(C_i|\mathcal{D})$) derivation}

Eq.~\ref{eq:p_ci_D_deriv} shows the derivation of $P(C_i|\mathcal{D})$. First, we introduced the hidden variable $H$, then replace $\mathcal{D}$ with $d_i$ as $C_i$ only depends upon $d_i$ in $\mathcal{D}$. The next, we apply Bayes rule to $P(C_i|d_i,H)$ and drop $H$ from $P(d_i|C_i,H)$, as $d_i$ does not depend upon $H$ when $C_i$ is given (Figure~\ref{fig:annotators_graph_model}). Finally, it replace $P(d_i|H)$ with $\sum_{C_i}P(d_i|C_i)P(C_i|H)$.

\begin{equation} \label{eq:p_ci_D_deriv}
\begin{split}
    P(C_i|\mathcal{D}) &= \sum_{H} P(C_i, H|\mathcal{D})\\
    &= \sum_{H} P(C_i|d_i, H) P(H|\mathcal{D}) \\
    &= \sum_{H} \frac{P(d_i|C_i, H) P(C_i|H)}{P(d_i|H)} P(H|\mathcal{D}) \\
    &= \sum_{H} \frac{P(d_i|C_i) P(C_i|H)}{P(d_i|H)} P(H|\mathcal{D}) \\
    &= \sum_{H} \frac{P(d_i|C_i) P(C_i|H)}{\sum_{C_i} P(d_i|C_i)P(C_i|H)} P(H|\mathcal{D}),
\end{split}
\end{equation}

\acrodef{MAP}{maximum a posteriori}

\end{document}